%% file: main.tex
\documentclass[conference]{IEEEtran}
\IEEEoverridecommandlockouts
\usepackage{cite}
\usepackage{amsmath,amssymb,amsfonts}
\usepackage{algorithmic}
\usepackage{graphicx}
\usepackage{textcomp}
\usepackage{xcolor}
\def\BibTeX{{\rm B\kern-.05em{\sc i\kern-.025em b}\kern-.08em
    T\kern-.1667em\lower.7ex\hbox{E}\kern-.125emX}}
\usepackage{algorithm}
\usepackage{algorithmic}

\usepackage{booktabs} % for professional tables
\usepackage{multirow} 

\usepackage{bbm}
\usepackage{amsthm}   % For theorems, definitions, etc.
\usepackage{tcolorbox}
\usepackage{url}

\usepackage{enumitem}

\usepackage{listings}
\usepackage{courier}
\theoremstyle{definition}

\usepackage{xspace}

\newcommand{\jasper}{\textsc{Jasper}\xspace}

\newcommand{\assertLLM}{\textsc{AssertLLM}\xspace}

\newcommand{\kgmodel}{\textsc{AssertionForge}\xspace}
\newcommand{\numData}{five\xspace}

\newcommand{\apb}{\textsc{apb}\xspace}
\newcommand{\ethmac}{\textsc{ethmac}\xspace}
\newcommand{\omsp}{\textsc{openMSP430}\xspace}

\newcommand{\uart}{\textsc{uart}\xspace}
\newcommand{\sockit}{\textsc{sockit}\xspace}

\newcommand{\KGnoRTL}{\textsc{AssertionForge w/o RTL}\xspace}
\newcommand{\baselineKG}{\textsc{AssertionForge Orig KG}\xspace}

\usepackage[colorinlistoftodos]{todonotes}

\definecolor{applegreen}{rgb}{0.55, 0.71, 0.0}

\begin{document}

\title{AssertionForge: Enhancing Formal Verification Assertion Generation with Structured Representation of Specifications and RTL\\
}

\author{
Yunsheng Bai, Ghaith Bany Hamad, Syed Suhaib, Haoxing Ren \\
NVIDIA \\
\texttt{yunshengb, gbanyhamad, ssuhaib, haoxingr@nvidia.com}
}

\maketitle

\input{sec-abstract.tex}


\input{sec-intro.tex}
\input{sec-related.tex}

\input{sec-model.tex}
\input{sec-result.tex}

\input{sec-conc.tex}

{
% \small
\bibliographystyle{plain}
\bibliography{bibliography}
}

\appendices

\input{sec-impl.tex}
\input{sec-eff.tex}

\input{sec-more-exp.tex}

\end{document}

%% file: sec-abstract.tex
\begin{abstract}

Generating SystemVerilog Assertions (SVAs) from natural language specifications remains a major challenge in formal verification (FV) due to the inherent ambiguity and incompleteness of specifications. Existing LLM-based approaches, such as \assertLLM, focus on extracting information solely from specification documents, often failing to capture essential internal signal interactions and design details present in the RTL code, leading to incomplete or incorrect assertions. We propose a novel approach that constructs a Knowledge Graph (KG) from both specifications and RTL, using a hardware-specific schema with domain-specific entity and relation types. We create an initial KG from the specification and then systematically fuse it with information extracted from the RTL code, resulting in a unified, comprehensive KG. This combined representation enables a more thorough understanding of the design and allows for a multi-resolution context synthesis process which is designed to extract diverse verification contexts from the KG. Experiments on \numData designs demonstrate that our method significantly enhances SVA quality over prior methods. This structured representation not only improves FV but also paves the way for future research in tasks like code generation and design understanding.

% \yba{TODO: mention even if spec is relatively lengthy and complete, augmenting the initial KG with RTL can still be helpful to further annotate/check/resolve consistency...}

% Existing LLM-based approaches, such as AssertLLM, primarily focus on extracting information solely from specification documents, often failing to capture essential internal signal interactions and design details embedded within the RTL code. This reliance on incomplete or potentially ambiguous specifications, coupled with the limitations of purely textual analysis in capturing intricate behavioral and structural nuances, can lead to inaccuracies and gaps in the generated assertions.

% We present \amodel, an end-to-end framework for automating the generation of Formal Verification test plans and SystemVerilog Assertions from natural language design specifications. Leveraging LLMs and self-optimizing knowledge graphs, \amodel bridges the gap between design intent and low-level verification, improving productivity and quality. It extracts information from specifications, builds knowledge graphs, aligns terms with signals, generates test plans, and translates them into SVAs. The system is refined via a feedback loop incorporating \jasper results and reference test plans. We evaluate \amodel using an open-source benchmark dataset and demonstrate its effectiveness in automating FV, reducing manual effort, and improving test plan accuracy. This research advances the application of LLMs in hardware verification and has the potential to transform industry practices.

\end{abstract}

%% file: sec-intro.tex
\section{Introduction} %\YS{the introduction is too long. compress the DSA part.}
\label{sec-intro}

Formal Verification (FV) is a critical process in the VLSI design flow, ensuring that increasingly complex hardware designs conform to their specifications~\cite{clarke2018model}. Among FV techniques, Assertion-Based Verification (ABV) uses SystemVerilog Assertions (SVAs), which are formal properties that describe expected design behavior, to verify functional correctness~\cite{vijayaraghavan2005practical}. To be effective, these assertions must encapsulate both the high-level design intent, as described (often ambiguously) in natural language specifications~\cite{foster2008assertion}, and the low-level implementation details codified in Register-Transfer Level (RTL) code. Verification engineers traditionally bridge this gap by mentally synthesizing information from both sources, a process that is difficult to automate yet essential for producing high-quality SVAs—assertions that are syntactically correct, functionally accurate, and provide comprehensive coverage of critical design properties.

Recently, Large Language Models (LLMs) have shown promise in generating SVAs from natural language specifications, thanks to their advanced text comprehension and coding capabilities. Methods like \assertLLM~\cite{fang2024assertllm} decompose the SVA generation task into phases, using customized LLMs to extract structural specifications, map signal definitions, and generate assertions by prompting the LLM with contextual information from specifications about relevant signals. These approaches demonstrate the feasibility of using LLMs for generating SVAs, but they remain fundamentally limited in their ability to capture the full design context.

The limitations stem from their approach to utilizing sources of verification information. Specification-focused approaches like \assertLLM extract design intent from the often ambiguous and incomplete specifications~\cite{foster2008assertion,alshazly2014detecting,Darbari2022,DBLP:journals/corr/abs-2004-05853,sanghvi2014specification}, and thus miss crucial implementation details found exclusively in the RTL, leading to plausible but functionally incorrect assertions. Conversely, RTL-focused approaches can capture implementation details but lack understanding of original design intent and high-level functionality, producing assertions that verify implementation without validating adherence to specifications. 

We hypothesize that explicitly constructing a structured, interconnected mental model of the design—one that links design intent with RTL behavior—will provide a more robust foundation for assertion generation. This mirrors how verification engineers manually synthesize information from multiple sources to write meaningful assertions. Compared to unstructured retrieval-based approaches~\cite{fang2024assertllm,lewis2020retrieval}. such a mental model enables more precise tracing of intent to implementation and facilitates deeper reasoning, allowing the LLM to identify signal dependencies and complex interactions. However, building such a model is challenging, requiring structured knowledge extraction from specifications, analysis of RTL control flows, and alignment between abstract requirements and implementation details.

In this paper, we present \kgmodel, a novel approach that constructs a unified \textbf{Knowledge Graph (KG)} from specifications and RTL, thereby actively ``forging'' assertions by bridging the semantic gap between high-level specifications and RTL implementation, enabling discovery of complex multi-signal interactions often obscured in linear text. The KG is built in two stages: entity-relation extraction from the specification via an LLM and structural parsing of RTL via a Verilog parser.
% This unified KG is constructed in two stages: first, a customized GraphRAG~\cite{edge2024local} framework uses an LLM to extract entities and relationships from the specification, guided by a domain-specific schema tailored for hardware verification. Second, a Verilog parser extracts information from the RTL, enriching the KG with implementation details and establishing cross-domain links.

With the unified KG constructed, the next step is crafting effective prompts for LLMs to generate high-quality test plans and SVAs. We propose a \textbf{multi-resolution context synthesis} process that progressively narrows focus from broad design overviews to fine-grained signal interactions: (1) LLM-based global summarization provides a high-level view of the specification and RTL; (2) Signal-specific retriever extracts relevant snippets from the specification and RTL; (3) \textbf{Guided Random Walk with Adaptive Sampling} (GRW-AS) explores the KG, tracing structured paths that connect the target signal to critical design elements. To ensure concise yet informative prompts, an LLM-based pruning mechanism filters and refines the retrieved text and graph-derived paths, balancing diversity and relevance before guiding final SVA generation.

Our contributions are summarized as follows:
\begin{itemize}
    \item We propose a novel KG-based approach, \kgmodel, that integrates information from both natural language specifications and RTL code into a unified structured representation for SVA generation.
    % \item We introduce a hardware-specific schema for the KG, featuring domain-specific entity and relation types that capture the nuanced behavioral and structural aspects of hardware designs.
    \item We develop a multi-resolution context synthesis process, which constructs verification-critical prompts by providing context as global summaries, signal-specific snippets, and structured paths from the KG.
    \item We demonstrate that our method significantly enhances SVA quality through experiments on \numData designs, achieving superior results compared to existing LLM-based methods. 
    % We are among the first to adopt a coverage-based metric to evaluate SVA generation for FV.
    % \item We outline potential future applications of our approach in other tasks, such as code generation and design understanding, paving the way for broader use cases in the field of hardware design automation.
\end{itemize}

%% file: sec-related.tex
\section{Related Work} %\YS{the introduction is too long. compress the DSA part.}
\label{sec-related}

\subsection{Formal Verification and SVA Generation}

% LLM-based approaches for SystemVerilog Assertion (SVA) generation have gained traction, showing promising results in processing full specification documents~\cite{orenes2023using,mali2024chiraag,FVEval2024,ayalasomayajula2024lasp,maddala2024laag,kande2024security,akyash2024self,liu2024domain,kumar2024generative}. \assertLLM~\cite{fang2024assertllm} automates assertion extraction from specifications, while \cite{harris2016glast} and \cite{zhao2019automatic} explore formal grammar learning and subtree analysis for assertion synthesis. Beyond LLMs, methods like GOLDMINE~\cite{vasudevan2010goldmine}, HARM~\cite{germiniani2022harm}, A-Team~\cite{danese2017team}, and AutoSVA~\cite{orenes2021autosva} apply data mining, static analysis, and templates respectively for assertion generation. However, these approaches often struggle with complex specifications, suffer from limited coverage, or depend heavily on simulation stimuli quality. Recent works also focus on RTL-guided SVA generation, with \cite{kande2023llm} targeting security properties and \cite{sun2023towards} leveraging circuit-aware translation to improve assertion quality.

LLM-based approaches for SystemVerilog Assertion (SVA) generation have gained traction~\cite{orenes2023using,mali2024chiraag,kang2024fveval,ayalasomayajula2024lasp,maddala2024laag,kande2024security,akyash2024self,liu2024domain,kumar2024generative,pulavarthi2025llmsreadypracticaladoption}, with \assertLLM~\cite{fang2024assertllm} automating assertion extraction from specifications. Non-LLM methods include GOLDMINE~\cite{vasudevan2010goldmine}, HARM~\cite{germiniani2022harm}, A-Team~\cite{danese2017team}, and AutoSVA~\cite{orenes2021autosva}, which use data mining, static analysis, templates, etc. for assertion generation. However, these approaches often struggle with complex specifications, suffer from limited coverage, or depend heavily on simulation stimuli quality. Recent RTL-guided SVA works include \cite{kande2023llm} and \cite{sun2023towards}. Our work focuses on advancing LLM-based SVA generation, exploring how structured representation and context-aware prompting can improve assertion quality.

\subsection{LLM for Knowledge Graph Construction and Understanding}

Traditional methods for Knowledge Graph (KG)~\cite{ji2021survey,abu2021domainKG} construction rely on non-LLM approaches requiring manual effort and domain-specific ontologies~\cite{martinez2018openie,bosselut2019comet,ye2022generative,zhong2023comprehensive}. Recent advancements leverage LLMs for KG construction, automatically inferring entities and relationships~\cite{trajanoska2023enhancing,pan2024unifying,bi2024codekgc,zhu2024llms}. LLM-based graphs and KGs have gained popularity across domains~\cite{kau2024combining,peng2024graph}, with Microsoft's GraphRAG demonstrating their application in question answering~\cite{edge2024local}. KGs enhance LLM-based NLP tasks including question answering, dialogue systems, and reasoning~\cite{wang2024knowledge,ren2024explicit,sarmah2024hybridrag,liang2024survey,wang2024cross} by improving knowledge-grounded responses and contextual understanding. Our approach adapts these advancements to hardware verification by integrating specifications and RTL code into a knowledge-rich graph.

%% file: sec-model.tex
\section{Methodology} 
\label{sec-model}

% \subsection{Overview}

\begin{figure*}[h]
\centering
\includegraphics[width=0.87\textwidth]{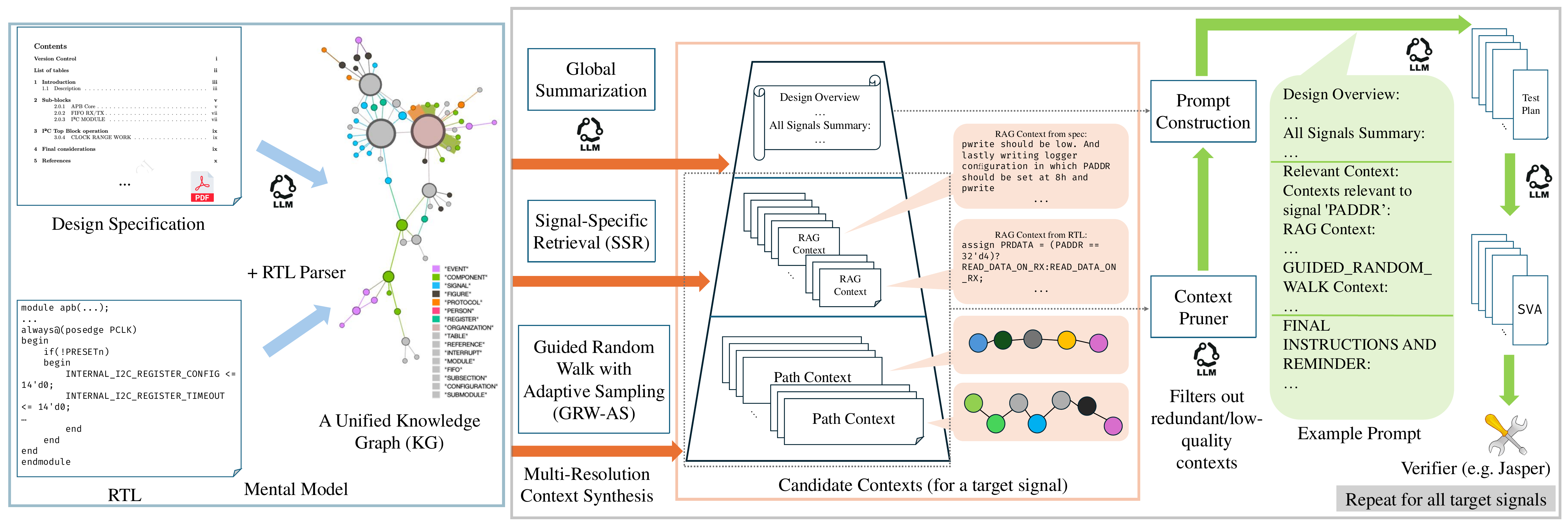}
\vspace{-0.3cm}
\caption{Overview of \kgmodel. Our method is structured into a three-stage process: First, we construct a domain-specific Knowledge Graph (KG) that captures the essential entities and relationships in the design. Together with the raw specifications and RTL, we build a mental model for the design under verification. Second, we generate three types of candidate contexts for each signal. Lastly, we use these contexts to dynamically construct prompts for a Large Language Model (LLM) to generate accurate SVAs. 
% This section describes the problem setup, the KG construction and refinement process, and the dynamic prompt construction methodology.
}
\vspace{-0.3cm}
\label{fig:model_architecture}
\end{figure*}

In this section, we introduce our novel approach, \kgmodel (Figure~\ref{fig:model_architecture}), for generating high-quality SystemVerilog Assertions (SVAs) from both natural language specifications and RTL code. 

\subsection{Problem Setup and Overview of \kgmodel}

We formalize the problem as follows: Given a hardware design specification $\mathcal{S}$ and RTL code $\mathcal{R}$, we generate SystemVerilog Assertions $\mathcal{A} = \{A_1, \ldots, A_n\}$ evaluated for syntactic correctness and functional accuracy. To emulate verification engineers' cognitive process~\cite{foster2008assertion} — a workflow rooted in cognitive science's concept of \textbf{mental model} (internal representations humans construct to understand complex systems~\cite{johnson1983mental}) — we develop a computational approach integrating $\mathcal{S}$ (design intent) and $\mathcal{R}$ (implementation details).

To computationally realize this mental model, we construct a Knowledge Graph (KG), $\mathcal{G} = (\mathcal{V}, \mathcal{E})$, where $\mathcal{V}$ represents nodes (entities) and $\mathcal{E}$ represents edges (relationships). Entities $\mathcal{V}$ are derived from both $\mathcal{S}$ and $\mathcal{R}$, and edges $\mathcal{E}$ capture relationships defined by our domain-specific schema. Formally, $\mathcal{G} = \mathcal{F}(\mathcal{S}, \mathcal{R})$, where $\mathcal{F}$ is the KG generation function. In our approach, the complete mental model comprises both this structured KG representation alongside the raw specification and RTL texts, enabling context-aware SVA generation through dynamically constructed LLM prompts.

\subsection{Knowledge Graph Construction with GraphRAG}
\label{sec:kg-construction}
GraphRAG~\cite{edge2024local} is a framework for generating knowledge graphs from textual documents through a process of \textbf{chunking} source text, \textbf{LLM-based extraction} of entities and relationships, \textbf{summarization} of node/edge descriptions, and \textbf{graph assembly}. While GraphRAG's default extraction focuses on general-purpose entities, hardware verification demands specialized entity and relationship types.

We propose an effective procedure to customize GraphRAG for hardware verification. First, we leverage \textbf{automated LLM-based schema discovery} by presenting a small representative set of hardware specification excerpts to the LLM, prompting it to identify domain-specific entity types (e.g., ``registers'', ``clocks'') and relation types (e.g., ``triggers'', ``connects\_to''). This is followed by \textbf{schema review}, where we manually refine the LLM-generated schema by merging redundant or overlapping concepts (e.g., combining ``Reset\_Signal'' and ``Reset''). Finally, through \textbf{prompt update and indexing}, we modify only GraphRAG's entity extraction prompt to incorporate our hardware-specific schema while keeping all other GraphRAG indexing components unchanged.

The schema $\Sigma$ is formally defined as a tuple $(\mathcal{E}_t, \mathcal{R}_t)$ consisting of \textbf{entity types} $\mathcal{E}_t$ representing key hardware design components (e.g., ``design specification'', ``module'', ``signal'', ``port'', ``register'', ``clock'') and \textbf{relation types} $\mathcal{R}_t$ representing relationships (e.g., ``hasSection'', ``contains'', ``defines'', ``connectsTo'', ``operatesAt'').

Our customized entity extraction prompt ensures that nodes and edges in $\mathcal{G}_0$ contain detailed attributes such as type, description, source ID, and relationships. \enspace Node attributes include fields like entity type, description, level, and degree, while edge attributes capture weight, description, and source IDs of connected nodes. These attributes are crucial for accurately representing both high-level and low-level design aspects. We will publicly release the exact prompt used for entity extraction and our entire source code to enable reproducibility.

\subsection{Refinement with RTL Information}
\label{subsec:rtl-refine}

To enhance the completeness and accuracy of the KG, we further refine $\mathcal{G}_0$ using information extracted from the RTL code $\mathcal{R}$. We utilize a specialized Verilog parser based on \cite{Takamaeda:2015:ARC:Pyverilog} to extract detailed information from RTL design files, including modules, ports, signals, control flow structures, and signal assignments. This process, denoted as $\mathcal{G} = \psi(\mathcal{G}_0, \mathcal{R})$, analyzes Verilog files to augment the initial KG with RTL-specific elements. New nodes are added for internal signals, control flow constructs, and module instances, while existing nodes are enriched with attributes like signal width and type information. Edges represent both structural relationships (e.g., ``has\_port'') and behavioral relationships (e.g., ``assigns\_to''). Our analysis also identifies finite state machines by recognizing clocked always blocks with case statements, creating FSM nodes that capture state transition logic.

Correspondences between specification and RTL nodes are identified via string matching of signal and module names. We use exact matching for direct references and a fuzzy matching algorithm to handle minor variations in naming conventions. This linking process establishes traceability between requirements and implementation.
% The resulting refined graph $\mathcal{G}$ typically contains over 1,000 nodes and 1,500 edges for even moderately sized designs, capturing both the structure and behavior of the hardware design at multiple levels of abstraction.

\subsection{Multi-Resolution Context Synthesis for SVA Generation}
\label{subsec:mrcs}

Current approaches to LLM-based hardware verification predominantly rely on Retrieval-Augmented Generation (RAG)~\cite{fang2024assertllm,lewis2020retrieval}. While RAG can effectively retrieve documentation snippets and code fragments using signal names as queries, the retrieved context often contains noise, and important behavioral patterns may be scattered across multiple documents and not captured in a single retrieval. 
% critical multi-signal interactions are difficult to discover through text-based retrieval alone.

To address these challenges, we propose a dynamic context synthesis framework that unifies multiple context generators with an intelligent LLM-based pruning mechanism. Our framework consists of a suite of context generators
% \[
% \Gamma = \{\gamma_1, \gamma_2, \ldots, \gamma_M\},
% \]
where each generator leverages distinct heuristics to capture different aspects of the design:
% For instance, the \textbf{RAG Context Retriever} varies retrieval parameters (e.g., chunk size, similarity thresholds) to produce diverse text-based contexts from both specifications and RTL. In contrast, the \textbf{Graph Context Retriever} extracts structured subgraphs from the Knowledge Graph to capture multi-signal interactions. This retriever identifies:
\begin{enumerate}[label=(\roman*)]
\item \textbf{Global Summarization:} (broadest level) An LLM-based summarizer extracts high-level design intent from specifications and RTL, forming a cohesive system-level overview.

% \item \textbf{Motif-based contexts:} We assume that recurring structural patterns in hardware designs map to specific functional behaviors that should be verified. Our implementation detects three key motif types: (1) \emph{star motifs}, where multiple signals converge on a central controller; (2) \emph{handshake patterns}, characterized by request-acknowledge signal pairs; and (3) \emph{pipeline stages}, where data validity signals propagate through successive processing stages. We utilize the VF2 algorithm~\cite{cordella2004sub} for subgraph pattern matching to detect these structures.

\item \textbf{Signal-Specific Retrieval (SSR):} Given a target architectural signal, a RAG-based retriever extracts localized, functionally relevant snippets from the specification and RTL, ensuring precise contextual grounding.

\item \textbf{Guided Random Walk with Adaptive Sampling (GRW-AS):} Given a node in the KG corresponding to a target signal, a novel graph-based traversal algorithm (detailed in Section~\ref{sec:grw-as}) explores the KG, generating paths that connect the target signal to other key design elements.

\end{enumerate}

% These graph-based heuristics are designed to reveal structural relationships that are often missed by conventional text retrieval.

Since naively combining all contexts would overwhelm downstream processing, we introduce an \emph{LLM-as-Pruner}. This pruner evaluates each candidate context generated by SSR and GRW-AS — evaluating relevance, diversity, and verification potential — and filters out redundant or low-quality contexts. 
% Formally, let \(P: \mathcal{C} \rightarrow \{0,1\}\) be a binary function where \(P(c) = 1\) indicates context selection. The pruned set is:
% \[
% \mathcal{C}^* = \{ c \in \mathcal{C} \mid P(c) = 1 \}.
% \]
The pruner considers both the individual merit of each context and its complementarity with other selected contexts to ensure comprehensive coverage of verification properties. Due to space constraints, we omit the full LLM-pruner prompt; however, it will be released upon paper acceptance.

The pruned contexts are then packed into dynamic prompts—up to the maximum token limit—to provide the LLM with comprehensive yet focused information. This ensures that the LLM does not just process more data, but processes better-selected data. By dynamically adapting context selection based on verification needs, our approach enables the LLM to generate test plans and SystemVerilog Assertions (SVAs) that capture complex behavioral patterns that are missed by traditional RAG-only methods.

% Our approach performs a two-stage generation pipeline. First, the LLM uses the enriched contexts to generate natural language (NL) test plans that capture verification intentions for the target signals. These plans serve as an intermediate representation of verification goals. In the second stage, these test plans guide the LLM in generating SystemVerilog Assertions (SVAs) that precisely capture the complex behavioral patterns identified in the first stage. 

Following \assertLLM, we define architectural signals as input/output ports and architectural-level registers explicitly mentioned in the specification, excluding internal signals further implemented in RTL. For each architectural signal, we construct a dedicated prompt by combining the global summaries with the LLM-pruned contexts from SSR and GRW-AS. These contexts are packed across multiple prompts, each varying in content but constrained by the LLM’s maximum token limit, ensuring diverse yet focused verification inputs. Each enriched prompt is then used to generate natural language (NL) test plans and actual SystemVerilog Assertions (SVAs). The generated test plans and SVAs for all architectural signals are combined for final evaluation.

% \subsection{Guided Random Walk with Adaptive Sampling (GRW-AS)}
% \label{sec:grw-as}

% These factors are combined to compute transition probabilities at each step, adaptively guiding the walk. Walks terminate upon reaching another interface signal or exceeding a predefined step budget. 

\subsection{Guided Random Walk with Adaptive Sampling (GRW-AS)}
\label{sec:grw-as}

To efficiently explore the KG's structural and behavioral relationships, we introduce the Guided Random Walk with Adaptive Sampling (GRW-AS) algorithm. Inspired by biased random walk techniques~\cite{weiss1983random} used in recommendation systems~\cite{cooper2014random}, knowledge discovery~\cite{backstrom2011supervised}, network analysis~\cite{costa2007exploring}, and graph representation learning~\cite{grover2016node2vec,hamilton2017inductive}, GRW-AS performs targeted exploration of the design, prioritizing paths likely to be relevant for verification. Starting from the target architectural signal's corresponding node, GRW-AS performs multiple, guided random walks through the KG. Each walk is biased by a combination of:

\begin{enumerate}[label=(\roman*)]
    \item \textbf{Local Node Importance}: Prioritizing structurally significant nodes (e.g., modules, key registers), thus guiding the walk toward verification-critical regions of the KG. The importance score is computed as $I(n) = 0.4 \cdot \frac{deg(n)}{max\_degree} + 0.6 \cdot T(n)$, where $deg(n)$ is the degree of node $n$, $max\_degree$ is the maximum degree of any node in the KG, and $T(n)$ represents a mapping from a node type to its semantic significance which can be manually adjusted using domain knowledge of the design under verification. In our current implementation, we use a uniform weight for all nodes.
    
    \item \textbf{Directional Guidance}: Favoring paths that lead towards other known architectural signals. In the initialization phase, we compute shortest paths between all pairs of architectural signals and store the ``next hop'' for each signal-to-signal path. For a candidate node $c$ during the walk, the direction score is calculated as $D(c) = \frac{\sum_{t \in \text{architectural\_signals}} \mathbb{I}(c \text{ is next hop to } t)}{|\text{architectural\_signals}|}$, where the indicator function equals 1 if $c$ is the first node along the shortest path from the current node to target architectural signal $t$. This effectively biases the walk towards nodes that serve as ``stepping stones'' to reach other architectural signals.
        
    \item \textbf{Exploration novelty}: Encouraging exploration of less-visited areas of the graph by prioritizing unvisited nodes. Quantified as $N(c) = \begin{cases} 1.0 & \text{if } c \notin visited \\ 0.0 & \text{otherwise} \end{cases}$, where $visited$ is the set of already visited nodes in the current walk, which is dynamically updated during each walk, making the sampling adaptive.
\end{enumerate}

These factors are combined to compute transition probabilities at each step according to the formula:
$P(c) = \alpha \cdot I(c) + \beta \cdot D(c) + \gamma \cdot N(c)$
where $\alpha$, $\beta$, and $\gamma$ are configurable weights controlling the influence of each factor. The probabilities are normalized to sum to 1 before node selection. Walks terminate either upon reaching another architectural signal or exceeding a predefined step budget. 
% Additionally, a teleportation mechanism with probability 0.1 is incorporated to allow occasional jumps to "gateway nodes" that frequently appear in paths between interface signals, helping to escape local regions of the graph.

%% file: sec-result.tex
\section{Experiments}

We evaluate our approach, \kgmodel, on \numData designs. 
% Following existing work, we assess the quality of generated SystemVerilog Assertions (SVAs) using syntax correctness and pass/fail criteria. Additionally, we evaluate the coverage of proven SVAs, considering only those assertions that are proven (i.e., no counterexample according to the Jasper tool). 
All of the datasets and model implementation will be released upon paper acceptance.

% \yba{TODO: mention we use gpt-4 as LLM backend. Also describe baselines. Also describe Jasper tool version and details with citation.}

% \yba{TODO: write a subsection on dataset details like where we obtain these 5 designs (source), etc. Sort of like a table neded with columns: design name, design source (OpenCore etc. maybe giving a url link...), design brief description, NL spec length *(pages, tokens), RTL \# files, \# tokens, etc.}

\subsection{Dataset and Knowledge Graph Statistics}

We evaluate our approach on \numData diverse hardware designs. These designs represent a range of complexities and functionalities in hardware systems, from communication protocols to microcontrollers and cryptographic modules. For each design, our focus is on generating SVAs for a specific module within the design. Since Verilog allows modules to reference or import other modules, we consider the total number of RTL files associated with each design, as shown in Table \ref{tab:dataset}. 
% The table also provides an overview of these designs, including their sources, brief descriptions, and key statistics related to their specifications and RTL implementations.
% Table \ref{tab:dataset} also presents the statistics of the Knowledge Graphs constructed for each design. The table shows the number of nodes and edges in the RTL refined KG, as well as the time taken to construct the KG via GraphRAG. 
% These results demonstrate that the KG construction scales with the complexity of the design, with larger designs such as \ethmac and \omsp taking longer to process and yielding larger KGs. Even for complex designs, the construction time remains practical, enabling timely integration into the SVA generation workflow.

\begin{table*}[t]
\centering
\footnotesize
\caption{Overview of the hardware designs and Knowledge Graph statistics used in our evaluation. Token counts are obtained using the cl100k\_base tokenizer\cite{openai2023tokenizer}. Time refers to the KG construction time in minutes.}
\label{tab:dataset}
\begin{tabular}{lccccccccc}
\hline
Design & Source & Description & Spec Pages & Spec Tokens & RTL Files & RTL Tokens & \# Nodes & \# Edges & Time \\
\hline
\apb & \cite{opencores2024} & APB to I2C interface & 12 & 2,634 & 1 & 1,116 & 128 & 225 & 4.25 \\
\ethmac & \cite{opencores2024} & Ethernet MAC transmit control & 102 & 33,754 & 4 & 6,262 & 1332 & 2502 & 8.10 \\
\omsp & \cite{fang2024assertllm} & MSP430 microcontroller & 129 & 34,133 & 29 & 116,093 & 4636 & 5800 & 6.67 \\
% \tinyPairing & \cite{fang2024assertllm} & Tate bilinear pairing core & 17 & 3,674 & 8 & 15,179 & 210 & 171 & 1.41 \\
\sockit & \cite{fang2024assertllm} & FPGA-based 1-wire master & 29 & 13,102 & 4 & 11,547 & 678 & 1400 & 4.41 \\

\uart & \cite{fang2024assertllm} & UART to Bus interface & 10 & 3,103 & 6 & 9,235 & 1490 & 1815 & 1.60 \\

\hline
\end{tabular}
\end{table*}

\begin{table*}[t]
\centering
\caption{Evaluation results. 
         (\#SVA: total assertions generated; \#SynC: syntactically correct; \#Proven: proven or passing;
          COI Statement/Branch/Functional/Toggle: coverage metrics for each coverage model under COI.)}
\label{tab:results}
\begin{tabular}{lccccccc}
\hline
\multirow{2}{*}{Model} & \multirow{2}{*}{\#SVA} & \multirow{2}{*}{\#SynC} & \multirow{2}{*}{\#Proven} 
  & \multicolumn{4}{c}{COI Coverage (\%)} \\
\cline{5-8}
 & & & & Statement & Branch & Functional & Toggle \\
\hline
\multicolumn{8}{c}{\apb} \\
\hline
\assertLLM & 221	 & 170	 & 56 & 100.00	& 86.67	& 84.79	& 84.16 \\
\KGnoRTL & 188	& 128	& 22 & 100.00	& 86.67	& 70.05	& 68.32 \\
\baselineKG & 551	& 412 &	92 & \textbf{100.00}	& \textbf{100.00}	& \textbf{100.00} &	\textbf{100.00} \\
\kgmodel & \textbf{615} &	\textbf{549}	 & \textbf{220} & \textbf{100.00}	& \textbf{100.00}	& \textbf{100.00} &	\textbf{100.00} \\
\hline

\multicolumn{8}{c}{\ethmac} \\
\hline
\assertLLM & 520	 & 106	 & 15 & 45.10	 & 44.83	 & 49.01	 & 50.29 \\
\KGnoRTL & 1345	 & 331	 & 42 & 50.98	 & 51.72 & 	71.21	& 77.46 \\
\baselineKG & 1365 & 	428	 & 74 & 94.12 & 	93.10	 & 77.80	 & 72.83 \\
\kgmodel & \textbf{1673} & \textbf{960} & \textbf{208} & \textbf{100.00}	 & \textbf{100.00}	 & \textbf{99.12}	 & \textbf{98.84} \\
\hline

\multicolumn{8}{c}{\omsp} \\
\hline
\assertLLM & 409	 & 128 & 	55 & 94.70 & 	95.05	 & 87.32 & 	85.81\\
\KGnoRTL & 1123	 & 216	 & 106 & 99.08	& 98.91	 & 88.94 &	86.95 \\
\baselineKG & 1202 & 	323	 & 153  & 	99.54	 & 99.52	 & 89.87	 & 87.96 \\
\kgmodel & \textbf{1600}	 & \textbf{698} & \textbf{327}  & \textbf{100.00} & 	\textbf{99.76} & \textbf{90.09} & \textbf{88.16} \\
\hline

\multicolumn{8}{c}{\sockit} \\
\hline
\assertLLM & 118	& 49	& 13 & 85.71	& 82.35	& 88.54 & 	90.46 \\
\KGnoRTL & 393	& 131	& 41 & \textbf{100.00}	& \textbf{100.00}	& 99.33 &	99.08 \\
\baselineKG & 429	& \textbf{156}	& 32 & \textbf{100.00}	& \textbf{100.00}	& \textbf{99.78}	& \textbf{99.69} \\
\kgmodel & \textbf{448}	& 136	& \textbf{49} & \textbf{100.00}	& \textbf{100.00}	& \textbf{99.78}	& \textbf{99.69} \\
\hline

\multicolumn{8}{c}{\uart} \\
\hline
\assertLLM & 249 & 	83	 & 29 & 74.36  &	78.26  & 	66.78	 & 63.01 \\
\KGnoRTL & 233	 & 93	 & 28  & 74.36	& 78.26	& 70.72	& 68.49 \\
\baselineKG & \textbf{279}	& \textbf{174} &	\textbf{50} & \textbf{84.62}	 & \textbf{84.78}	& 83.88	& 83.56 \\
\kgmodel & 253	 & 132	 & 27 & \textbf{84.62} & \textbf{84.78} & \textbf{88.82} & \textbf{90.41} \\
\hline

\end{tabular}
\end{table*}

\subsection{Baselines and Implementation Details}

For all methods, 
% including the baselines and our proposed \kgmodel, 
we use GPT-4o~\cite{openai2023gpt4} as the LLM backend.
% for both KG construction and SVA generation. 
KG construction follows the GraphRAG framework~\cite{edge2024local} with default settings, customizing only the prompt for entity and relation extraction. RTL parsing utilizes `pyverilog'~\cite{Takamaeda:2015:ARC:Pyverilog}. For SSR, we adopt a multi-scale hierarchical chunking strategy following \cite{lewis2020retrieval,borgeaud2022improving}, with chunk sizes of 50, 100, 200, 800, and 3200 tokens and overlap ratios of 0.2 and 0.4. We retrieve the top 20 ranked chunks per signal-name query using a hybrid similarity model combining TF-IDF~\cite{salton1988term} and Sentence Transformers~\cite{reimers2019sentence}. The GRW-AS algorithm runs 70 walks per signal with a budget of 100 steps, using $\alpha=0.3$, $\beta=0.5$, and $\gamma=0.2$. The LLM-based pruner retains a maximum of 50 contexts per type and 100 in total. 

% After pruning, contexts from different sources (global summaries, SSR, and GRW-AS) are grouped by type and distributed across multiple prompts while adhering to the GPT-4o's token limit—allocating up to 75\% for context and reserving 25\% for the LLM’s response. Each method is given a fixed per-signal prompt budget (set to 3) to ensure a fair comparison in SVA generation. This constraint simulates real-world verification scenarios where engineers, limited by time and resources, iteratively refine prompts by exploring different contexts or phrasings to improve assertion quality.

To ensure a fair comparison and simulate a realistic verification scenario, we limit each method to a maximum of 3 prompts per architectural signal. Global summaries and pruned contexts from SSR and GRW-AS are grouped by type, and then distributed across these prompts, maximizing context inclusion while respecting the GPT-4o's token limit per prompt.

We compare \kgmodel against three baselines: (1) \textbf{\assertLLM} \cite{fang2024assertllm}: An LLM-based approach that generates SVAs directly from natural language specifications without constructing a KG. (2) \textbf{\KGnoRTL}: This baseline is similar to our full \kgmodel, but it does not perform RTL refinement, relying solely on the initial KG $\mathcal{G}_0$ constructed from the design specification. (3) \textbf{\baselineKG}: This baseline uses the full \kgmodel pipeline but uses the vanilla GraphRAG prompt for KG construction instead of our domain-customized prompt with hardware-specific schema and examples.

% This setup allows us to assess the impact of KG construction and RTL refinement on the quality of generated SVAs. 

\subsection{Evaluation Protocol}

We use Cadence \jasper{} (version 2023.12) to evaluate our generated SystemVerilog Assertions (SVAs).  We measure the total number of SVAs generated (\textbf{\#SVA}), the number of syntactically correct SVAs (\textbf{\#SynC}), and the number of SVAs proven without counterexamples (\textbf{\#Proven}).  Assuming a golden RTL implementation, this work focuses on generating high-quality SVAs; debugging counterexamples to determine root cause (RTL bug versus assertion error) is left for future work.  Coverage is evaluated only for \emph{proven} assertions using Jasper's \emph{cone-of-influence (COI)} coverage~\cite{chockler2003coverage,li2019coverage} models. We report \textbf{COI Functional Coverage}, \textbf{COI Branch Coverage}, \textbf{COI Statement Coverage}, and \textbf{COI Toggle Coverage}, measuring coverage of high-level functionality, conditional branches, statements, and signal transitions, respectively, within the assertion's influence. COI coverage is obtained using the \texttt{check\_cov} command in Jasper.

\begin{table*}[h]
\centering
\caption{Ablation study results on the \uart{} design, showing the impact of removing individual components from our framework.}
\label{tab:ablation_uart}
\begin{tabular}{lcccccccc}
\hline
\multirow{2}{*}{Model Variant} & \multirow{2}{*}{\#SVA} & \multirow{2}{*}{\#SynC} & \multirow{2}{*}{\#Proven} 
  & \multicolumn{4}{c}{COI Coverage (\%)} \\
\cline{5-8}
 & & & & Statement & Branch & Functional & Toggle \\
\hline
\textbf{Full \kgmodel} (all components) & 253 & 132 & 27 & 84.62 & 84.78 & 88.82 & 90.41 \\
No Summaries (w/o global summarization) & 236 &  76 & 31 & 84.62 & 84.78 & 78.29 & 75.80 \\
No RAG (w/o retrieval-augmented generation) & 311 & 128 & 38 & 84.62 & 84.78 & 83.55 & 83.11 \\
No GRW-AS (w/o guided random walk) & 243 & 115 & 26 & 84.62 & 84.78 & 83.55 & 83.11 \\
No Pruner (w/o LLM-based context pruning) & 281 & 144 & 20 & 84.62 & 84.78 & 83.88 & 83.56 \\
Spec-RTL Concatenation &  91 &  36 &  9 & 74.36 & 78.26 & 70.39 & 68.04 \\
\hline
\end{tabular}

\vspace{-0.2cm}

\end{table*}

\subsection{Results and Analysis}

Table~\ref{tab:results} presents the evaluation results across \numData hardware designs, where \kgmodel consistently outperforms the baselines in both proven assertions and coverage. The gains are especially notable for more complex designs, achieving near-perfect coverage across multiple metrics. Although \baselineKG generally surpasses \assertLLM, it still falls short of the full \kgmodel pipeline, underscoring the impact of our domain-specific KG construction. These results highlight not just the quantity but the quality of our generated assertions — a critical distinction since plausible but incorrect assertions waste computational resources.

\subsection{Ablation Study}

To analyze the contribution of each component in \kgmodel, we conduct an ablation study on the \uart design as in Table~\ref{tab:ablation_uart}. Removing the \emph{guided random walk algorithm} (No GRW) significantly reduces the number of proven assertions, highlighting its importance in exploring relevant design paths. The \emph{global summarization component} (No Summaries) drastically reduces syntactically correct assertions while producing mixed effects on coverage metrics. 
% Disabling the \emph{retrieval-augmented generation} (No RAG) or the \emph{LLM-based context pruner} (No Pruner) leads to moderate performance changes across metrics. 
The baseline approach that simply concatenates specification and RTL text (Spec-RTL Concat) performs worst across all metrics, demonstrating that naively combining all available information overwhelms the LLM and is impractical for industry-scale designs with substantially larger specifications and RTL.

\subsection{Case Study: Knowledge Graph Visualization}

\begin{figure}[h]
\centering
\includegraphics[width=0.47\textwidth]{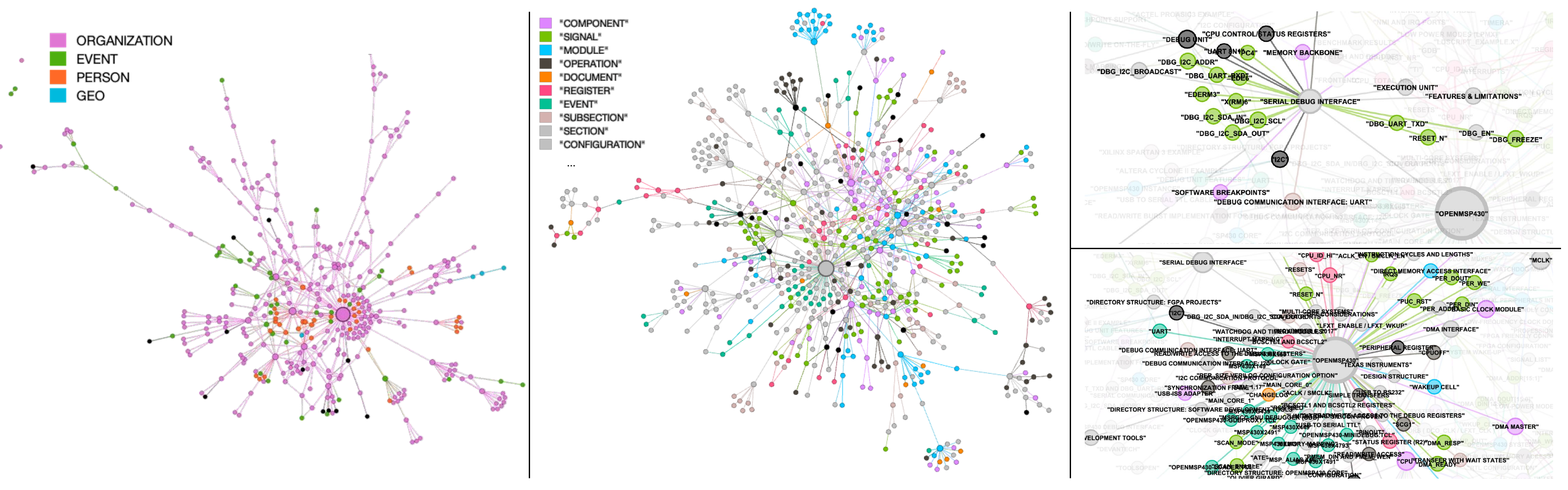}

\caption{Visualization of the KGs from \omsp{} using \cite{ICWSM09154}. Node colors represent different types of entities, such as modules and signals. Node colors indicate the \textit{type} attribute of each entity. Left: KG generated via the original entity extraction prompt of GraphRAG. Middle: KG generated via our domain-customized prompt. Right: Two zoomed-in views of the KG, highlighting key entities with their labels. 
% The figures are generated via the Gephi software~\cite{ICWSM09154}.
}
% \vspace{-0.5cm}
\label{fig:kg_vis_openmsp}
\end{figure}

Figure~\ref{fig:kg_vis_openmsp} visualizes KGs constructed from the OPENMSP430 design specification, contrasting the impact of our domain-specific schema.
% The left image shows a KG generated with the vanilla GraphRAG prompt.
% , resulting in few, generic node types. 
% The center and right images demonstrate our domain-specific KG.
% , revealing a richer structure with hardware-relevant entities and relationships. 
% This highlights the increased detail and complexity captured by our approach, even prior to RTL refinement.

\subsection{Case Study: SVAs Generated for \ethmac}

We illustrate that \kgmodel{} generates high-quality SVAs capturing essential design properties on the \ethmac{} design in Figure~\ref{fig:ethmac_svas}, an Ethernet MAC transmit control module.  The first SVA ensures `ByteCnt' resets and `TxFlow' halts during reset, which is crucial for maintaining transmission integrity, while the second verifies the correct assertion of `WillSendControlFrame' during the transmission of control frames, ensuring proper data transmission.

\begin{figure}[htbp]
  \centering
  \begin{tcolorbox}[
      colback=white,
      colframe=black,
      boxrule=0.5pt,
      left=5pt,
      right=5pt,
      top=5pt,
      bottom=5pt,
      width=\columnwidth,
      title=Selected SVAs Generated for \ethmac{},
      fonttitle=\bfseries,
      coltitle=black,
      colbacktitle=white
  ]
  \footnotesize
  \textbf{Plan:} Validate that when `ResetByteCnt' and `xReset' are active, `ByteCnt' is set to zero and `TxFlow' halts within the next clock cycle.  
  \textbf{SVA:} 
\begin{verbatim}
@(posedge MTxClk) 
  (ResetByteCnt == 1 && TxReset == 1) 
  |-> (ByteCnt == 0 && TxFlow == 0);
\end{verbatim}
  % \medskip
  \textbf{Plan:} Ensure that `WillSendControlFrame' is asserted just before `TxStartFrmIn' is asserted and remains asserted until `TxEndFrmIn' is asserted.  
  \textbf{SVA:}
\begin{verbatim}
@(posedge MTxClk) 
  ((WillSendControlFrame && TxStartFrmIn) 
   && !TxEndFrmIn) 
  |-> WillSendControlFrame;
\end{verbatim}
  \end{tcolorbox}
  \caption{Examples of SVAs generated by \kgmodel.}
  \label{fig:ethmac_svas}
  \vspace{-0.4cm}
\end{figure}

\subsection{Case Study: Paths Extracted by GRW-AS Algorithm}
To illustrate the effectiveness of our GRW-AS algorithm, we examine key paths extracted from two designs' KGs that may have contributed to generating high-quality SVAs. Due to space constraints, we show selected subpaths that capture the essential signal relationships.
% — full paths in our KG contain significantly more nodes.

\textbf{Path 1: \uart Transmission Control.} This path reveals critical dependencies:
\begin{small}
\begin{verbatim}
tx_busy (port) → uart_tx (module) → 
uart_top (module) → new_tx_data (port) → 
data_stability (verification_point)
\end{verbatim}
\end{small}
\noindent Together with global summaries and signal-specific context, \kgmodel synthesized this path information to generate this key SVA:
\begin{small}
\begin{verbatim}
@(posedge clock) (!tx_busy && new_tx_data) 
    |-> ##1 tx_busy
\end{verbatim}
\end{small}

\textbf{Path 2: \apb Interrupt Generation.} This subpath exposes interrupt control flow:
\begin{small}
\begin{verbatim}
INT_TX (assignment) → apb (module) → 
TX_EMPTY (port) → data_stability (point) → 
reset_behavior (point) → PRESETn (port)
\end{verbatim}
\end{small}
\noindent These KG-derived insights, combined with specification snippets and global design context, led to critical SVAs for FIFO status and reset behavior:
\begin{small}
\begin{verbatim}
@(posedge PCLK) TX_EMPTY |=> INT_TX
@(posedge PCLK) (!PRESETn) |-> TX_EMPTY
\end{verbatim}
\end{small}

%% file: sec-conc.tex
\section{Conclusion and Future Work}
\label{sec-conc}

We introduced \kgmodel, a novel approach for generating SystemVerilog Assertions (SVAs) that integrates Knowledge Graph (KG) and Register-Transfer Level (RTL) information. Experimental results on \numData designs demonstrate significant improvements over baseline methods.
% , particularly for complex designs produces higher-quality assertions.
% more assertions with higher syntax correctness and provability rates, as well as improved functional coverage.
% These results highlight the effectiveness of combining high-level specifications with low-level implementation details in a structured KG representation for SVA generation. 
While our approach shows promising results, future work could explore more sophisticated KG construction techniques, methods to further improve functional coverage, and extensions to other hardware design automation tasks such as automatic code generation and design error detection. 
% Overall, \kgmodel represents a significant step towards more efficient and effective automated formal verification in hardware design.

%% file: sec-impl.tex
% First appendix section
\section{Implementation Details}
\label{app:impl}

\subsection{Overall Process}
Algorithm~\ref{alg:main-process} outlines the core workflow of \kgmodel, integrating specification analysis, RTL parsing, and multi-resolution context synthesis.

\begin{algorithm}[htbp]
\small
\caption{\kgmodel Main Process}
\label{alg:main-process}
\begin{algorithmic}[1]
\REQUIRE Design specification $\mathcal{S}$, RTL code $\mathcal{R}$
\ENSURE Generated SVAs $\mathcal{A}$

\STATE Parse specification: $\mathcal{S} \leftarrow \text{PDFParser}(\mathcal{S})$
\STATE Build initial KG: $\mathcal{G}_0 \leftarrow \mathcal{F}(\mathcal{S})$ \COMMENT{Sec.~\ref{sec:kg-construction}}
\STATE Refine KG with RTL: $\mathcal{G} \leftarrow \psi(\mathcal{G}_0, \mathcal{R})$ \COMMENT{Sec.~\ref{subsec:rtl-refine}}
\STATE Extract valid architectural signals: $\mathcal{V} \leftarrow \text{RTLSignalExtractor}(\mathcal{R})$
\STATE Generate global summary: $\mathcal{C}_g \leftarrow \text{DesignSummarizer}(\mathcal{S}, \mathcal{R}, \mathcal{V})$

\FOR{\textbf{each} signal $v_i \in \mathcal{V}$}
    \STATE Retrieve signal context: $\mathcal{C}_r \leftarrow \text{SSR}(v_i, \mathcal{S}, \mathcal{R})$
    \STATE Extract KG paths: $\mathcal{C}_k \leftarrow \text{GRW-AS}(\mathcal{G}, v_i, \mathcal{V})$ \COMMENT{Sec.~\ref{sec:grw-as}}
    \STATE Prune contexts: $\mathcal{C}_p \leftarrow \mathcal{P}(\mathcal{C}_g \cup \mathcal{C}_r \cup \mathcal{C}_k)$
    
    \STATE Initialize prompt budget: $\mathcal{B} \leftarrow 3$ \COMMENT{Max prompts per signal}
    
    \FOR{$j \leftarrow 1$ \textbf{to} $\mathcal{B}$}
        \STATE Construct dynamic prompt: 
        $\mathcal{D}_{i,j} \leftarrow \Omega(\mathcal{C}_p, v_i, j)$ \COMMENT{Sec.~\ref{subsec:mrcs}}
        \STATE Generate NL plans: $\mathcal{P}_{i,j} \leftarrow \mathcal{L}_{\text{plan}}(\mathcal{D}_{i,j})$
        \STATE Synthesize SVAs: $A_{i,j} \leftarrow \mathcal{L}_{\text{sva}}(\mathcal{P}_{i,j}, \mathcal{D}_{i,j})$
        \STATE Verify assertions: $\text{Jasper}(A_{i,j})$
    \ENDFOR
\ENDFOR

\STATE Aggregate coverage metrics: $\text{AnalyzeResults}(\bigcup_{i,j} A_{i,j})$
\end{algorithmic}
\end{algorithm}

\subsection{Knowledge Graph Construction from Specifications}
\label{app:kg-construction}

The creation of a comprehensive Knowledge Graph (KG) from design specifications is a critical first step in our approach. We leverage a customized version of GraphRAG~\cite{edge2024local} for this purpose, adapting it specifically for hardware verification domain knowledge. 
The entity extraction process is guided by a detailed prompt that defines the domain-specific schema (Figure~\ref{fig:entity-extraction-prompt}). This prompt is crucial for ensuring that the LLM extracts relevant entities and relationships from the hardware specifications.

\begin{figure}[htbp]
\centering
\begin{tcolorbox}[
    colback=white,
    colframe=black,
    boxrule=0.5pt,
    left=5pt,
    right=5pt,
    top=5pt,
    bottom=5pt,
    width=\columnwidth,
    title=Prompt for Entity Extraction,
    fonttitle=\bfseries,
    coltitle=black,
    colbacktitle=white
]
\footnotesize
[The following content is summarized or omitted for brevity:
- Goal, Steps, and Additional instructions for hardware design specifications.
For the full prompt, please refer to the GraphRAG repository.]

\textbf{Entity Types:}

Design Specification, Section, Subsection, Table, Figure, Author, Module, Submodule, Protocol, Signal, Port, Register, FIFO, Clock, Interrupt, Operation, Frequency, Standard, Reference, Component, Version, Date, Comment, Pin, Configuration, Constraint/Rule, Address, Document, Block

\textbf{Relation Types:}

hasSection, hasSubsection, contains, authoredBy, defines, implements, uses, connectsTo, configures, generatesInterrupt, operatesAt, compliesWith, referencedIn, inputTo, outputFrom, partOf, interfacesWith, triggersOperation, dependsOn, transmitsData, receivesData, has\_input, has\_output, has\_register, performs, uses\_protocol, has\_constraint, has\_rule, describes, references, modifies, creates, closes/fixes, HasSubModule, HasSignal, HasPort, UsesProtocol, UsesClock, UsesAddress, DescribesOperation, TriggersInterrupt, RevisionHistory, belongsToSection, illustrates, hasAuthor, has\_port, connected\_to, described\_in

\textbf{Examples:}

[Example outputs and entity-relationship formats demonstrated in multiple contexts, including hardware design specifications and narrative texts.]

\textbf{-Real Data-}

Entity\_types: \{entity\_types\}  

Text: \{input\_text\}

\end{tcolorbox}
\caption{Example prompt for entity extraction (abbreviated). For the full prompt, see: 
\texttt{https://github.com/microsoft/graphrag/blob/main/} \texttt{graphrag/index/graph/extractors/graph/prompts.py}}
\label{fig:entity-extraction-prompt}
\end{figure}

A key advantage of our approach is the adaptability of the knowledge graph schema. Formal verification engineers can modify the entity and relation types based on their domain knowledge and the specific design under verification. For example, when verifying a processor design, additional entity types like "Pipeline Stage," "Execution Unit," or "Branch Predictor" could be added to the schema. Similarly, when verifying a memory controller, types like "Memory Channel," "Timing Parameter," or "Refresh Cycle" might be more relevant.

\subsection{RTL Parsing and Knowledge Graph Refinement}
\label{subsec:rtl-refine}

The initial knowledge graph, $\mathcal{G}_0$, constructed from the design specification, is refined by integrating structural and behavioral information extracted from the RTL code, $\mathcal{R}$. This process, denoted as $\mathcal{G} = \psi(\mathcal{G}_0, \mathcal{R})$, is crucial for bridging the gap between high-level design intent and low-level implementation details. We achieve this using a custom RTL analysis pipeline built upon the PyVerilog library~\cite{Takamaeda:2015:ARC:Pyverilog}. Algorithm~\ref{alg:rtl-parsing} provides an overview of the RTL parsing procedure.

The RTL parsing and knowledge graph refinement were implemented using Python 3.10.14 and PyVerilog version 1.3.0.

\begin{algorithm}[htbp]
\small
\caption{RTL Parsing and KG Refinement}
\label{alg:rtl-parsing}
\begin{algorithmic}[1]
\REQUIRE Initial KG $\mathcal{G}_0$, RTL code $\mathcal{R}$
\ENSURE Refined KG $\mathcal{G}$

\STATE $\mathcal{G} \leftarrow \mathcal{G}_0$ \COMMENT{Initialize with specification KG}
\STATE $\mathcal{R} \leftarrow \text{preprocess\_includes}(\mathcal{R})$ \COMMENT{Inline included files}
\STATE $\text{rtl\_elements} \leftarrow \text{parse\_rtl}(\mathcal{R})$ \COMMENT{Extract modules, ports, signals, etc.}
\STATE $\text{dataflow} \leftarrow \text{analyze\_dataflow}(\mathcal{R})$ \COMMENT{Extract signal dependencies}

\FOR{\textbf{each} module $m \in \text{rtl\_elements.modules}$}
    \STATE Add module node to $\mathcal{G}$
    \STATE Add nodes for module ports, signals, instances, FSMs
    \STATE Add edges for structural relationships (containment, connections)
    \STATE Add nodes and edges for behavioral relationships (assignments, control flow)
\ENDFOR

\STATE Link specification nodes in $\mathcal{G}_0$ to corresponding RTL nodes in $\mathcal{G}$
\STATE Add root node connecting all components if needed for graph connectivity

\RETURN $\mathcal{G}$
\end{algorithmic}
\end{algorithm}

The parsing process begins by preprocessing the RTL files to handle include directives, inlining the contents of included files to create a self-contained representation. We extract key structural and behavioral information through several specialized methods.

For structural information, we leverage PyVerilog's parser to extract hierarchical component information. Module definitions are identified by traversing the Abstract Syntax Tree (AST) and locating ``ModuleDef'' nodes. For each module, we extract port declarations (``Input'', ``Output'', ``Inout'' nodes) from the AST, capturing direction (input/output/inout) and width details (using the ``Width'' node in the AST). This process determines whether a port is an input or output and extracts its bit width through direct AST analysis. Module instantiations are identified within the parent module's definition via ``Instance'' nodes, along with their port connections by examining the ``portlist'' attribute.

For behavioral information, we leverage a multi-strategy approach to identify FSMs. The primary method examines always blocks with clock sensitivity lists that contain case statements, a common FSM implementation pattern. We supplement this with pattern matching to identify state variables, looking for signals named with common patterns like "state", "st\_", or "current". For instance, when we encounter code patterns like "always @(posedge clk)" followed by "case(current\_state)", we recognize a potential FSM. Within these ``always'' blocks, we look for ``case'' statements (represented by ``Case'' nodes in the AST). We supplement this AST-based approach with pattern matching on signal and parameter names, looking for common FSM-related identifiers like ``state'', ``current\_state'', ``next\_state'', ``st\_'', or ``fsm''.  This helps identify FSMs even when coding styles deviate from the strict ``always @(posedge clk)'' and ``case(state)'' pattern.

Control flow structures are another critical aspect of RTL behavior. We extract conditional statements (if/else), case statements, and loops using a combination of AST traversal and pattern matching. For example, a regex pattern identifies case statement conditions, which often represent state-dependent behavior. Similarly, we analyze assignments, distinguishing between continuous assignments (assign statements) and procedural assignments in always blocks.

Signal assignments are extracted from both ``Assign'' nodes (for continuous assignments) and from within ``always'' blocks (for procedural assignments).  For each assignment, the left-hand side (LHS) signal (the target of the assignment) and the right-hand side (RHS) expression are recorded.  The RHS expression is further analyzed to identify the signals involved, establishing dependencies. We distinguish between blocking and non-blocking assignments, which is crucial for correctly modeling Verilog semantics.

PyVerilog's ``VerilogDataflowAnalyzer'' is used to construct a dataflow graph. The analyzer's ``getBindings'' method is used to obtain the dataflow bindings for each signal, revealing its dependencies. We add this dataflow information directly to our knowledge graph by creating edges between signal nodes.

After extracting this RTL information, we refine the knowledge graph as shown in Algorithm~\ref{alg:rtl-parsing}. New nodes are added for RTL elements, including modules, ports, signals, module instances, FSMs, control flow structures, and assignments. Existing nodes from the initial specification KG ($\mathcal{G}_0$) are preserved.  Edges are created to represent:

\begin{enumerate}[label=(\roman*)]
    \item \textbf{Containment:} Modules contain ports, signals, instances, FSMs, and control structures.
    \item \textbf{Instantiation:} A module instantiates another module.
    \item \textbf{Port Connections:} Signals are connected to ports in module instances.
    \item \textbf{Dataflow:} Signals drive other signals through assignments.
    \item \textbf{Control Flow:} Control structures (if, case, loops) influence signals/assignments.
    \item \textbf{FSM Structure:} An edge goes from FSM to module.
    \item \textbf{Assignment:} An edge goes from assignment to lhs and rhs.
\end{enumerate}

Finally, a linking process, detailed in Section~\ref{subsubsec:fuzzy-matching}, establishes connections between nodes in the specification KG ($\mathcal{G}_0$) and the RTL-derived KG ($\mathcal{G}$). The result is a unified KG representing both design intent and implementation details.

\subsection{Fuzzy Signal Name Matching}
\label{subsubsec:fuzzy-matching}

A critical aspect of bridging the specification and RTL domains is establishing correspondence between entity names that may differ due to naming conventions or abbreviations. Our approach leverages a combination of exact and fuzzy matching techniques to link entities between the specification and RTL knowledge graphs.

For exact matching, we identify direct name references in the specification text using regular expressions, seeking word boundary-anchored occurrences of signal and module names from the RTL. However, hardware specifications often refer to signals using variants of their RTL names. For example, a specification might mention "reset signal" while the RTL uses "rst\_n" or "PRESETn".

To address this, we implement a specialized fuzzy matching algorithm that considers:

\begin{enumerate}[label=(\roman*)]
    \item \textbf{Common hardware abbreviations:} We maintain a dictionary mapping between full terms and common hardware abbreviations (e.g., "reset" → "rst", "clock" → "clk").
    
    \item \textbf{Case and separator variations:} We normalize names by converting to lowercase and removing separators like underscores, allowing matches between variants like "data\_valid" and "DataValid".
    
    \item \textbf{Active-low signal conventions:} We detect common patterns for active-low signals (e.g., "\_n", "\_b" suffixes or negation prefixes like "n" or "not\_").
    
    \item \textbf{Levenshtein distance:} For remaining candidates, we calculate edit distance with thresholds proportional to name length, capturing minor typographical variations.
\end{enumerate}

The matching process uses a scoring system where exact matches receive the highest score (1.0), followed by abbreviation expansions (0.9), case/separator variations (0.8), and edit-distance-based matches with scores decreasing as distance increases. We require a minimum match score of 0.6 and prioritize the highest-scoring match when multiple candidates exist.

For example, given an RTL signal "tx\_data\_valid", our algorithm would successfully match specification mentions of "transmit data valid", "TX\_DATA\_VALID", or "tx data valid signal". This fuzzy matching is particularly valuable for signals with domain-specific naming, where conventional string similarity metrics alone would fail to establish correct correspondence.

\subsection{RTL Signal Extractor}
\label{subsec:rtl-signal-extractor}

In the \kgmodel framework, the RTL Signal Extractor is essential for identifying and extracting valid signals from the RTL code. This process begins by parsing the RTL code to locate module declarations and their interfaces, typically encompassing \texttt{input}, \texttt{output}, and \texttt{inout} signals. Using regular expressions, the extractor identifies signal declarations, capturing details such as signal direction (\texttt{input}, \texttt{output}, \texttt{inout}), data type (\texttt{reg}, \texttt{wire}), and bit-width specifications. For instance, a signal declared as \texttt{input~[7:0]~data\_in} indicates an 8-bit wide input signal named \texttt{data\_in}. Once extracted, these signal names are compiled into a set of valid signals, serving as a reference for subsequent stages in the workflow as in Algorithm~\ref{alg:main-process}.

\subsection{Global Summarization}
\label{subsec:design-summarizer }

% The Design Summarizer (Global Summarization) synthesizes critical design context through structured prompts to large language models (LLMs). Four core prompts extract verification-relevant information from specifications and RTL code, formatted for maximal LLM comprehension.

The Design Summarizer (Global Summarization) component, denoted as $\mathcal{C}_g \leftarrow \text{DesignSummarizer}(\mathcal{S}, \mathcal{R}, \mathcal{V})$ in Algorithm~\ref{alg:main-process}, plays a pivotal role in distilling complex hardware design specifications and RTL code into concise, verification-focused summaries. This step is essential for providing the language model with a comprehensive understanding of the design context prior to assertion generation. Our implementation leverages a multi-faceted summarization approach that targets different aspects of the design through specialized prompts.

The Design Summarizer generates several complementary views of the design. First, it creates a high-level design specification summary (Figure~\ref{fig:design-summary-prompt}) that captures the main functionality and architectural components in 3-5 sentences. This is complemented by an RTL architecture summary (Figure~\ref{fig:rtl-summary-prompt}) that focuses on module hierarchy and interfaces. Additionally, it produces a comprehensive signals summary (Figure~\ref{fig:signals-summary-prompt}) that provides detailed information about each valid signal, including type, bit width, functionality, and interactions with other signals. Finally, it identifies key design patterns and protocols (Figure~\ref{fig:patterns-summary-prompt}) that have verification implications.

\begin{figure}[htbp]
\centering
\begin{tcolorbox}[
colback=white,
colframe=black,
boxrule=0.5pt,
left=5pt,
right=5pt,
top=5pt,
bottom=5pt,
width=\columnwidth,
title=Prompt for Design Specification Summary,
fonttitle=\bfseries,
coltitle=black,
colbacktitle=white
]
\footnotesize
You are an expert hardware design engineer. Please provide a concise summary (3-5 sentences)
of the following hardware design specification. Focus on the main functionality, key components,
and architecture. The summary should give a clear high-level understanding of what this design does.

Design Specification:
\{spec\_text\}

Provide only the summary, with no additional commentary or introduction.
\end{tcolorbox}
\caption{Prompt for generating a high-level design specification summary. The prompt emphasizes conciseness while focusing on functionality, key components, and architecture.}
\label{fig:design-summary-prompt}
\end{figure}

% For the RTL architecture summarization, we employ a similar approach but with emphasis on hierarchical structure and interfaces, as shown in Figure~\ref{fig:rtl-summary-prompt}.

\begin{figure}[htbp]
\centering
\begin{tcolorbox}[
colback=white,
colframe=black,
boxrule=0.5pt,
left=5pt,
right=5pt,
top=5pt,
bottom=5pt,
width=\columnwidth,
title=Prompt for RTL Architecture Summary,
fonttitle=\bfseries,
coltitle=black,
colbacktitle=white
]
\footnotesize
You are an expert hardware design engineer. Please provide a concise summary (3-5 sentences)
of the following RTL code. Focus on the module hierarchy, interfaces, and key architectural features.

RTL Code:
\{rtl\_text\}

Provide only the RTL architecture summary, with no additional commentary or introduction.
\end{tcolorbox}
\caption{Prompt for generating an RTL architecture summary, focusing on module hierarchy and interfaces.}
\label{fig:rtl-summary-prompt}
\end{figure}

% A more comprehensive analysis is performed for the signals summary, as detailed in Figure~\ref{fig:signals-summary-prompt}. This prompt is designed to extract detailed information about each signal's characteristics and relationships, providing critical context for assertion generation.

\begin{figure}[htbp]
\centering
\begin{tcolorbox}[
colback=white,
colframe=black,
boxrule=0.5pt,
left=5pt,
right=5pt,
top=5pt,
bottom=5pt,
width=\columnwidth,
title=Prompt for Comprehensive Signals Summary,
fonttitle=\bfseries,
coltitle=black,
colbacktitle=white
]
\footnotesize
You are an expert hardware verification engineer. Please analyze the following design specification and RTL code
to provide a comprehensive summary of the signals in the design. For each signal, include details about:

Signal name

Signal type (input, output, inout, internal, clock, reset, etc.)

Bit width (e.g., 1-bit, 8-bit, 32-bit)

Functionality and purpose

Key interactions with other signals

Valid Signals: {signals\_str}

Design Specification:
\{spec\_text\}

RTL Code:
\{rtl\_text\}

Focus on the signals listed above. If the RTL/spec doesn't provide information for a signal, make your best inference.
Format your response as a list with each signal having its own paragraph that includes all the details mentioned above.
Be concise yet complete.
\end{tcolorbox}
\caption{Prompt for generating a comprehensive summary of all signals in the design, including detailed technical characteristics.}
\label{fig:signals-summary-prompt}
\end{figure}

% The Design Summarizer also identifies verification-critical design patterns through a specialized prompt shown in Figure~\ref{fig:patterns-summary-prompt}.

\begin{figure}[htbp]
\centering
\begin{tcolorbox}[
colback=white,
colframe=black,
boxrule=0.5pt,
left=5pt,
right=5pt,
top=5pt,
bottom=5pt,
width=\columnwidth,
title=Prompt for Design Patterns Summary,
fonttitle=\bfseries,
coltitle=black,
colbacktitle=white
]
\footnotesize
You are an expert hardware design engineer. Please analyze the following design specification and RTL code
to identify and summarize key design patterns, protocols, or verification-critical structures.
Examples might include handshaking protocols, state machines, pipelines, arbiters, or clock domain crossings.

Design Specification:
\{spec\_text\}

RTL Code:
\{rtl\_text\}

Provide a concise summary (5-10 sentences) of the key design patterns and their verification implications.
\end{tcolorbox}
\caption{Prompt for identifying and summarizing key design patterns and protocols with verification implications.}
\label{fig:patterns-summary-prompt}
\end{figure}

Beyond these global summaries, the Design Summarizer also generates signal-specific descriptions for each target signal during the assertion generation process. This focused analysis, shown in Figure~\ref{fig:signal-specific-prompt}, provides detailed context about the specific signal for which assertions are being generated.

\begin{figure}[htbp]
\centering
\begin{tcolorbox}[
colback=white,
colframe=black,
boxrule=0.5pt,
left=5pt,
right=5pt,
top=5pt,
bottom=5pt,
width=\columnwidth,
title=Prompt for Signal-Specific Description,
fonttitle=\bfseries,
coltitle=black,
colbacktitle=white
]
\footnotesize
You are an expert hardware verification engineer. Please provide a detailed description of
the signal '\{signal\_name\}' based on the following specification and RTL.

Design Specification:
\{spec\_text\}

RTL Code:
\{rtl\_text\}

Include in your description:

The precise function of this signal
Its type (input, output, inout, internal, etc.) and bit width
Its timing characteristics (synchronous/asynchronous, edge-triggered, etc.)
Key relationships with other signals
How it affects or is affected by the overall system behavior
Any special conditions or corner cases related to this signal
Write 3-5 sentences with comprehensive, verification-focused details.
\end{tcolorbox}
\caption{Prompt for generating a detailed description of a specific signal, focusing on its function, characteristics, and interactions.}
\label{fig:signal-specific-prompt}
\end{figure}

The Design Summarizer implements an efficient caching mechanism to avoid redundant LLM calls. The global design summary is generated once and reused across all assertion generation tasks, while signal-specific descriptions are cached individually. This approach significantly reduces the overall LLM query cost while maintaining comprehensive context for assertion generation.

When constructing the final context for assertion generation, the summarizer combines these various summary components in a hierarchical manner, starting with the global design overview, followed by the RTL architecture summary, the target signal description, the comprehensive signals summary, and finally the design patterns summary. This layered approach provides the assertion generation model with both broad design context and signal-specific details, enabling it to generate more accurate and relevant SystemVerilog assertions.

\subsection{Signal-Specific Retrieval (SSR)}
\label{subsec:ssr}

The Signal-Specific Retrieval (SSR) component, represented as $\mathcal{C}_r \leftarrow \text{SSR}(v_i, \mathcal{S}, \mathcal{R})$ in Algorithm~\ref{alg:main-process}, is responsible for retrieving relevant contextual information from both specification documents and RTL code based on signal-specific queries. 

Our SSR implementation serves as a critical middle-resolution layer in our multi-level context synthesis framework, bridging the gap between the high-level global summarization and the fine-grained graph traversal of GRW-AS. While global summarization captures design-wide architectural intent and GRW-AS explores detailed signal-to-signal relationships, SSR provides essential functional context at the module and subsystem level. This intermediate resolution is crucial for hardware verification, as many behavioral specifications and protocol sequences exist precisely at this middle level of abstraction. 

To effectively capture this intermediate information, we adopt a multi-scale hierarchical chunking strategy inspired by the work of Lewis et al.~\cite{lewis2020retrieval} and Borgeaud et al.~\cite{borgeaud2022improving}, dividing documents into chunks of varying granularities that form what Robertson and Callan~\cite{robertson2009probabilistic} refer to as a "resolution pyramid." At the finest resolution (50 tokens), the system captures precise signal definitions and immediate interactions. The middle resolutions (100-200 tokens) capture functional behaviors and protocol sequences that typically span multiple sentences or paragraphs. The coarsest resolutions (800-3200 tokens) preserve architectural relationships and system-level behaviors that might be missed in smaller chunks.

To ensure coherent context across chunk boundaries, we implement overlapping chunks with ratios of 0.2 and 0.4, following the approach proposed by Dai and Callan~\cite{dai2019deeper}. This overlapping strategy addresses the "boundary fragmentation problem" identified by Hearst~\cite{hearst1997texttiling}, where critical information can be split across adjacent chunks. By creating strategic overlaps, we ensure that semantically coherent units (like descriptions of protocol sequences that might span multiple paragraphs) remain intact in at least some retrieved chunks.

Our retrieval framework uses two complementary methods for finding relevant context. First, we use TF-IDF to convert text chunks into sparse vectors that emphasize rare, distinguishing terms. Second, we use Sentence Transformers to generate dense semantic vectors that capture contextual meaning. For retrieval, we use the signal name (e.g., ``tx\_valid'' or ``data\_ready'') as the query term. We compute cosine similarity between this query and each chunk using both vector representations independently, then combine the scores with an average. This dual approach balances exact signal name matching with semantic relevance, allowing us to retrieve context even when signal names are referenced indirectly. 

To further refine the retrieval process, future enhancements could involve developing specialized embedding models fine-tuned on hardware design corpora. Such models would be adept at capturing the unique linguistic patterns and terminologies prevalent in hardware design documents, thereby improving retrieval accuracy. For instance, NV-Embed~\cite{lee2024nv}, a generalist embedding model, has demonstrated significant improvements in embedding tasks by incorporating architectural innovations and specialized training procedures. Adapting similar techniques to create embeddings tailored for hardware design verification could substantially enhance the performance of the SSR component.

\subsection{Guided Random Walk with Adaptive Sampling (GRW-AS)}  
\label{subsec:grw-as}

GRW-AS discovers semantically relevant paths through the knowledge graph using biased random walks guided by verification priorities. Algorithm~\ref{alg:grw-as} outlines the core procedure, implementing the three key biases defined in Section~\ref{sec:grw-as}.

\begin{algorithm}[htbp]
\small
\caption{GRW-AS Path Discovery}
\label{alg:grw-as}
\begin{algorithmic}[1]
\REQUIRE Knowledge graph $\mathcal{G}$, Start node $v_s$, signals $\mathcal{V}$  
\ENSURE Natural language path description $L$

\STATE $P \leftarrow [v_s]$, $v \leftarrow v_s$ \COMMENT{Initialize path}
\WHILE{step $\leq$ budget \AND $\mathcal{V} \neq \emptyset$}    
    \STATE $\mathcal{N} \gets \{u | (v,u) \in \mathcal{E}\}$ \COMMENT{Neighbor nodes}
    \STATE $\mathbf{P} \gets [\alpha I(u) + \beta D(u) + \gamma N(u)\ \forall u \in \mathcal{N}]$
    \STATE $v' \gets \text{sample}(\mathcal{N}, \mathbf{P})$ \COMMENT{Biased selection}
    \STATE $P$.append($v'$), $v \gets v'$
    \IF{$v' \in \mathcal{V}$}
        \STATE $\mathcal{V} \gets \mathcal{V} \setminus \{v'\}$
    \ENDIF
    \ENDWHILE
\STATE $L \gets \text{ConvertToNL}(P)$ \COMMENT{Natural Language (NL) description}
\RETURN $L$
\end{algorithmic}
\end{algorithm}

An example of a generated path description is shown in Figure~\ref{fig:grwas-path-example}. This example shows a path starting from the ``baud\_limit'' port in the \uart module, traversing through transmit and receive logic, and discovering critical signals through diverse relationship types. The description includes the type of each node (e.g., ``port'', ``module'', ``verification\_point'', ``protocol\_pattern'') and the module it belongs to, along with the varied relationships between consecutive nodes (e.g., ``drives'', ``input to'', ``involves'', ``includes''). While our current work focuses on representing paths as human-readable text, which we hypothesize is more natural for the LLM, future work could explore extracting subgraphs or motifs from the knowledge graph.

\begin{figure}[htbp]
\begin{tcolorbox}[
    colback=white,
    colframe=black,
    boxrule=0.5pt,
    top=1mm,
    bottom=1mm,
    left=1mm,
    right=1mm,
]
\begin{lstlisting}[
    basicstyle=\small\ttfamily,
    breaklines=true,
    columns=flexible
]
GUIDED RANDOM WALK FROM baud_limit (port)
Located in module: baud_gen
Path length: 57 nodes, discovered signals: 
tx_busy, ce_16, new_rx_data, rx_data, new_tx_data, baud_clk, baud_freq, clock

Signal flow path:
  baud_limit (port in baud_gen) drives baud_limit (port in uart_top)
  baud_limit (port in uart_top) input to uart_top (module)
  uart_top (module) part of new_tx_data_assignment (assignment in uart_top)
  new_tx_data_assignment (assignment in uart_top) assigns to new_tx_data (port in uart_top)
  new_tx_data (port in uart_top) involves data_stability (verification_point in uart_top)
  data_stability (verification_point in uart_top) involves baud_clk (port in uart_top)
  baud_clk (port in uart_top) includes rtl_111 (protocol_pattern in uart_top)
  rtl_111 (protocol_pattern in uart_top) found in uart_top (module)
  ...
  data_buf_assignment (assignment in uart_top) used in tx_data (port in uart_top)
  tx_data (port in uart_top) drives tx_data (port in uart_tx)
  tx_data (port in uart_tx) involves data_stability (verification_point in uart_tx)
  data_stability (verification_point in uart_tx) involves clock (port in uart_tx)
  clock (port in uart_tx) drives clock (port in uart_rx)
  ...
  rx_data (port in uart_rx) includes rtl_108 (protocol_pattern in uart_rx)
  rtl_108 (protocol_pattern in uart_rx) includes new_rx_data (port in uart_rx)
  ... [truncated]
\end{lstlisting}
\end{tcolorbox}
\caption{Example of a path description generated by GRW-AS.}
\label{fig:grwas-path-example}
\end{figure}

\subsection{LLM-based Context Pruner}
\label{subsec:llm-pruner}

The pruner analyzes each context's relevance, information density, and complementarity with other selected contexts, considering factors like explicit mentions of the target signal, descriptions of related signals, protocol specifications, and timing/behavioral constraints. As shown in Figure~\ref{fig:pruner-prompt}, the LLM is prompted to act as an expert verification engineer, receiving the original query, target signal name, and a set of contexts with metadata (source type, retrieval score, etc.). Crucially, the prompt instructs the LLM to select between a minimum and maximum number of contexts of the given type, explicitly encouraging the inclusion of even partially relevant information. The LLM outputs a list of indices indicating the selected contexts (e.g., ``Selected contexts: [0, 2, 5]''). Our implementation defaults to a maximum of 50 contexts per type and 100 total contexts, with a minimum of 2 contexts per type enforced whenever possible, promoting diversity.

\begin{figure}[htbp]
\centering
\begin{tcolorbox}[
colback=white,
colframe=black,
boxrule=0.5pt,
left=5pt,
right=5pt,
top=5pt,
bottom=5pt,
width=\columnwidth,
title=Prompt for LLM Context Pruner,
fonttitle=\bfseries,
coltitle=black,
colbacktitle=white
]
\footnotesize
You are an expert hardware verification engineer evaluating contexts to be used in generating verification plans for signal '\{signal\_name\}'.

QUERY: \{query\}

YOUR TASK: Select between \{min\_selection\} and \{max\_selection\} contexts of type '\{context\_type\}' that could help with verification.

IMPORTANT NOTES:

- Select at least \{min\_selection\} contexts even if they seem only indirectly relevant

- Consider both explicit mentions of '\{signal\_name\}' and general system information

- Partial information about interfaces, protocols, and system behavior is still valuable

- When in doubt, include rather than exclude contexts

CONTEXTS TO EVALUATE:

[CONTEXT 0]
\{context\_0\}

----

[CONTEXT 1]
\{context\_1\}

----

...

SELECTION INSTRUCTIONS:

1. SELECT AT LEAST \{min\_selection\} CONTEXTS, even if only partially relevant

2. Focus on contexts that might help verify \{signal\_name\}

3. Output your selection using ONLY the format "Selected contexts: [list of indices]"

For hardware verification, we need information about:

- Signal connections and dependencies

- Timing requirements

- Protocol details

- State transitions

- Interfaces

\end{tcolorbox}
\caption{Prompt template for the LLM-based Context Pruner, showing how contexts from different sources are presented for evaluation.}
\label{fig:pruner-prompt}
\end{figure}

\subsection{Multi-Resolution Context Synthesis}
\label{subsec:mrcs}

After pruning the contexts from various sources, the next critical step is generating effective verification plans and assertions through Multi-Resolution Context Synthesis. This process, denoted as $\mathcal{D}_{i,j} \leftarrow \Omega(\mathcal{C}_p, v_i, j)$ in Algorithm~\ref{alg:main-process}, dynamically constructs prompts that harmonize different context types across multiple resolution levels.

Our implementation assembles relevant contexts from SSR (middle resolution) and GRW-AS (fine-grained resolution), along with global design summaries (coarse resolution). For each signal, the system generates up to three distinct prompts ($\mathcal{B} = 3$ in our implementation), each containing complementary context combinations to maximize coverage of verification scenarios. These prompts respect a maximum token limit of 75\% of the LLM's context window, reserving space for the model's response.

The two-stage verification process begins with natural language (NL) plan generation, followed by SystemVerilog Assertion (SVA) synthesis. Figure~\ref{fig:nl-prompt} illustrates the prompt structure used for NL plan generation.

\begin{figure}[htbp]
\centering
\begin{tcolorbox}[
colback=white,
colframe=black,
boxrule=0.5pt,
left=5pt,
right=5pt,
top=5pt,
bottom=5pt,
width=\columnwidth,
title=Prompt for NL Test Plan Generation,
fonttitle=\bfseries,
coltitle=black,
colbacktitle=white
]
\footnotesize
Generate natural language test plans for signal '\{signal\_name\}'.

Relevant Context:
\{global\_summary\}

RAG Context:
\{rag\_context\}

GRW-AS Context:
\{grw\_context\}

CRITICAL - Valid Signal Names (USE ONLY THESE SIGNALS):
\{valid\_signals\}

Examples:

\{examples\}

Generate diverse test plans based on the given information. Each test plan should be on a new line and start with 'Plan: '.
\end{tcolorbox}
\caption{Prompt template for natural language verification plan generation, showing how multi-resolution contexts are integrated to guide the generation of signal-specific test plans.}
\label{fig:nl-prompt}
\end{figure}

After generating NL plans, the system synthesizes formal SVAs as shown in Figure~\ref{fig:sva-prompt}. This stage uses the same context synthesis approach but adds signal-specific context and in-context learning examples to guide correct assertion syntax.

\begin{figure}[htbp]
\centering
\begin{tcolorbox}[
colback=white,
colframe=black,
boxrule=0.5pt,
left=5pt,
right=5pt,
top=5pt,
bottom=5pt,
width=\columnwidth,
title=Prompt for SVA Generation,
fonttitle=\bfseries,
coltitle=black,
colbacktitle=white
]
\footnotesize
Generate SystemVerilog Assertions based on the following information:

Relevant Context:
\{global\_summary\}

\{signal\_specific\_summary\}

RAG Context:
\{rag\_context\}

GRW-AS Context:
\{grw\_context\}

Natural Language Test Plans for signal '\{signal\_name\}':

Plan 1: \{plan\_1\}

Plan 2: \{plan\_2\}

Plan 3: \{plan\_3\}

Examples:

\{examples\}

Generate one SVA for each of the provided natural language test plans. 
Enclose each SVA in triple backticks (```) and prefix it with 'SVA:'.
\end{tcolorbox}
\caption{Prompt template for SVA generation, including multi-resolution contexts, signal-specific natural language plans, and examples demonstrating the expected SVA format.}
\label{fig:sva-prompt}
\end{figure}

%% file: sec-eff.tex
\section{Efficiency Analysis}
\label{sec:efficiency}

We analyze \kgmodel's computational efficiency across the four designs. Table~\ref{tab:efficiency} summarizes the time distribution across major workflow stages, demonstrating our framework's practical scalability.

\begin{table*}[htbp]
\footnotesize
\centering
\caption{Time distribution across verification stages for different hardware designs}
\label{tab:efficiency}
\begin{tabular}{lccccc}
\hline
\textbf{Component} & \textbf{\apb} & \textbf{\ethmac} & \textbf{\omsp} & \textbf{\sockit} & \textbf{\uart} \\ \hline
KG Construction from Spec & 4.08\% & 10.65\% & 2.72\% & 0.62\% & 3.15\% \\
KG Refinement from RTL & 0.01\% & 0.07\% & 0.33\% & 0.01\% & 0.07\% \\
Context Summarization & 1.34\% & 0.00\% & 3.35\% & 1.17\% & 3.65\% \\
NL Plan Generation & 8.73\% & 5.12\% & 9.97\% & 3.97\% & 14.41\% \\
SVA Synthesis & 12.14\% & 5.83\% & 14.03\% & 4.48\% & 20.42\% \\
Jasper Verification & 73.26\% & 77.82\% & 69.31\% & 90.34\% & 57.00\% \\
Other Operations & 0.44\% & 0.51\% & 0.29\% & 0.41\% & 1.30\% \\ \hline
Total Runtime & 1h 47m & 1h 25m & 4h 12m & 4h 14m & 52m 20s \\ \hline
\end{tabular}
\end{table*}

Our analysis reveals that \kgmodel introduces reasonable computational overhead while enabling comprehensive verification. While the initial Knowledge Graph construction from specifications varies across designs (2.72-10.65\%), the subsequent KG refinement from RTL is remarkably efficient at less than 0.4\% of total runtime for all designs. 

The LLM-based components show good scalability. The dominant computational factor remains Jasper formal verification (57-78\% of runtime), which is an inherent cost in any verification approach. This verification phase could be parallelized in future implementations to further reduce overall time, though our current implementation evaluates each assertion sequentially.

It's worth noting that actual running times may vary based on factors such as the LLM backend engine speed, server load, and hardware configuration used for verification. The figures presented here represent typical performance observed in our experimental setup and are intended to illustrate the relative distribution of computational effort rather than absolute benchmarks. In practice, users may observe faster or slower performance depending on their specific environment and the optimizations applied to each component.

%% file: sec-more-exp.tex
\section{More Details on Experiments}

\subsection{Cone-of-Influence Coverage Metrics}

When measuring verification quality, we need to look beyond just counting assertions to examine how well they actually test the design's behavior. We chose Cadence \jasper's cone-of-influence (COI) coverage metrics as our quality indicator because they offer deeper insights into each assertion's effectiveness~\cite{chockler2003coverage}. 
% While traditional coverage simply reports how much of the design was exercised, COI coverage focuses on which specific design elements impact or are impacted by a given assertion, providing a more precise picture of verification thoroughness~\cite{li2019coverage}.

We use four complementary COI metrics to capture different verification aspects. \textbf{COI Statement Coverage} shows what percentage of RTL statements within the assertion's influence are tested, giving us basic code coverage. \textbf{COI Branch Coverage} reveals how well the assertion tests conditional paths in the logic, which is essential for verifying how the design makes decisions. \textbf{COI Functional Coverage} gauges whether the assertion truly verifies high-level requirements rather than just checking implementation details.

\textbf{COI Toggle Coverage} adds another dimension by tracking signal transitions within the assertion's reach, helping us find static logic that other metrics might overlook. These four metrics together give us a balanced view of assertion quality since each looks at different verification aspects. For example, an assertion might hit most statements but miss important branching logic, or it might check branches without exercising key signals.

\subsection{Impact of Prompt Count Parameter}
We conducted a parameter sensitivity study on the maximum prompt count parameter $\mathcal{B}$, which controls how many distinct prompts are generated per architectural signal. Table~\ref{tab:param_study} shows the results of varying $\mathcal{B}$ from our baseline value of 3 to higher values of 6 and 9 for the \uart design.

\begin{table*}[ht]
\centering
\caption{Effect of varying maximum prompt count parameter $\mathcal{B}$ on the \uart design.}
\label{tab:param_study}
\begin{tabular}{lcccccccc}
\hline
\multirow{2}{*}{$\mathcal{B}$ Value} & \multirow{2}{*}{\#SVA} & \multirow{2}{*}{\#SynC} & \multirow{2}{*}{\#Proven} 
  & \multicolumn{4}{c}{COI Coverage (\%)} \\
\cline{5-8}
 & & & & Statement & Branch & Functional & Toggle \\
\hline
$\mathcal{B} = 3$ (default) & 253 & 132 & 27 & 84.62 & 84.78 & 88.82 & 90.41 \\
$\mathcal{B} = 6$ & 394 & 255 & 53 & 84.62 & 84.78 & 83.93 & 83.64 \\
$\mathcal{B} = 9$ & 397 & 197 & 61 & 84.62 & 84.78 & 89.14 & 90.87 \\
\hline
\end{tabular}
\vspace{-0.2cm}
\end{table*}

Our results demonstrate that increasing the maximum prompt count $\mathcal{B}$ generally leads to more proven assertions, with the total count rising from 27 at $\mathcal{B}=3$ to 61 at $\mathcal{B}=9$. This improvement is expected, as each additional prompt contains complementary context combinations that can capture different verification scenarios. However, we observe that the relationship is not linear, with diminishing returns as $\mathcal{B}$ increases. 

% The fundamental reason for this behavior lies in the limited context window of the LLM. Each prompt must fit within this window, with our implementation reserving 75\% of the token limit for context and the remainder for the model's response. After applying the LLM-based context pruner, we distribute the remaining contexts across the allowed prompts, striving for balanced and diverse information in each. As $\mathcal{B}$ increases, we eventually reach a point where most verification-critical contexts have been covered in previous prompts, leading to diminishing returns from additional prompts.

Interestingly, we observe that coverage metrics do not consistently improve with increasing $\mathcal{B}$. For example, COI Functional Coverage actually decreases slightly from 88.82\% at $\mathcal{B}=3$ to 83.93\% at $\mathcal{B}=6$ before recovering to 89.14\% at $\mathcal{B}=9$. This non-monotonic behavior suggests that while more prompts generally lead to more assertions, the quality and diversity of those assertions depend on complex interactions between the contexts selected for each prompt.

In practice, the choice of $\mathcal{B}$ represents a trade-off between verification thoroughness and computational resources. Larger values yield more comprehensive verification but incur higher computational costs for both LLM inference and assertion verification. Based on our experiments, $\mathcal{B}=3$ offers a reasonable balance for practical applications, while higher values may be justified for critical components requiring exhaustive verification.

\subsection{Testbench SVAs Generated by LLM on the \omsp Design}

Figure~\ref{fig:osmp_svas} demonstrates SVAs generated by \kgmodel for the \omsp microcontroller, spanning clock domains, DMA behavior, interrupt handling, and debug interfaces. These automatically synthesized assertions exhibit three key characteristics: temporal correctness (sequencing requirements), protocol compliance (interface adherence), and data integrity (preventing unknown states).

\begin{figure*}[htbp]
\centering
\begin{tcolorbox}[
colback=white,
colframe=black,
boxrule=0.5pt,
left=5pt,
right=5pt,
top=5pt,
bottom=5pt,
width=\textwidth,
title=Selected SVAs for \omsp Generated by \kgmodel,
fonttitle=\bfseries,
coltitle=black,
colbacktitle=white
]
\footnotesize
\begin{minipage}[t]{0.48\textwidth}
\textbf{Clock Domain Validation:}
\begin{verbatim}
// A0: ACLK/SMCLK synchronization
@property p0
@(posedge smclk or aclk)
((smclk_en && !aclk_en) |-> (aclk == smclk));
\end{verbatim}

\textbf{Reset Behavior:}
\begin{verbatim}
// A22: Reset propagation
@property p22
@(posedge mclk) (puc_rst) |-> ##1 !aclk_en;
\end{verbatim}

\textbf{DMA Transaction Safety:}
\begin{verbatim}
// A109: Valid DMA input data
@property p109
@(posedge mclk) (dma_en && dma_we)
|-> (dma_din !== 'hX);
\end{verbatim}

\textbf{Interrupt Handling:}
\begin{verbatim}
// A180: IRQ-ACK consistency
@property p180
@(posedge mclk) (irq == 0) |-> irq_acc == 0;
\end{verbatim}

\textbf{Clock Gating:}
\begin{verbatim}
// A307: SMCLK stability
@property p307
@(posedge mclk) (smclk_en && cpu_en)
|-> $isunknown(smclk) == 0;
\end{verbatim}
\end{minipage}
\hfill
\begin{minipage}[t]{0.48\textwidth}
\textbf{Memory Access Validation:}
\begin{verbatim}
// A110: DMA address bounds
@property p110
@(posedge mclk) (dma_en &&
(dma_addr > 'hFFFF)) |-> (dma_resp == 1'b1);
\end{verbatim}

\textbf{Debug Interface Protocol:}
\begin{verbatim}
// A49: I²C SDA hold time
@property p49
@(posedge dbg_i2c_scl)
(dbg_en && dbg_i2c_sda_in |-> dbg_i2c_sda_out);
\end{verbatim}

\textbf{Power Management:}
\begin{verbatim}
// A23: DCO wakeup trigger
@property p23
@(posedge mclk)
(!dco_enable && cpu_en) |-> dco_wkup;
\end{verbatim}

\textbf{Asynchronous Event Handling:}
\begin{verbatim}
// A231: NMI to handler mapping
@property p231
@(posedge mclk)
disable iff (puc_rst) $rose(nmi) 
|-> $rose(cpu.NMI_handler);
\end{verbatim}

\textbf{Reset Assertion:}
\begin{verbatim}
// A283: Reset pin behavior
@property p283
@(posedge mclk)
(reset_n == 1'b0) |-> (puc_rst == 1'b1);
\end{verbatim}
\end{minipage}
\end{tcolorbox}
\caption{Representative SVAs for \omsp demonstrating \kgmodel's ability to capture diverse verification concerns..}
\label{fig:osmp_svas}
\vspace{-0.4cm}
\end{figure*}

These SVAs exhibit several features including multi-clock checks (A0), protocol timing requirements (A49), error propagation (A110), and cross-domain triggers (A23). This comprehensive coverage stems from the KG's unified representation of both specification requirements and RTL implementation details.

The guided random walks (GRW-AS) algorithm was particularly effective at discovering relationships between seemingly unrelated signals like ``cpu\_en'' and ``dco\_wkup'' by traversing the ``openMSP430.v'' connectivity paths. Additionally, examining the RTL implementation revealed critical behaviors not explicitly documented in the specification, such as the timing relationship between ``reset\_n'' and ``puc\_rst'' (A283) and the exact power management protocol between ``dco\_enable'' and ``dco\_wkup'' (A23). 

Despite the improvements in assertion quality and coverage achieved by \kgmodel, Table~\ref{tab:results} shows coverage is still not 100\% complete. Future work includes leveraging existing assertion libraries and learning from well-verified designs via RAG or in-context learning techniques. Additionally, iterative KG refinement, hardware-specific LLM fine-tuning, and implementing verification feedback loops could further enhance coverage.